\documentclass{ws-jmrr}
\usepackage[sort,compress,super]{cite}
\begin{document}

\catchline{0}{0}{2013}{}{}

\markboth{Patrick Brandao}{ Automatic polyp segmentation using convolution neural networks}

\title{Towards a computed-aided diagnosis system in colonoscopy: Automatic polyp segmentation using convolution neural networks}

\author{Patrick Brandao$^{a}$ 
, Odysseas Zisimopoulos$^a$, Evangelos  Mazomenos$^a$, Gastone Ciuti$^b$, Jorge Bernal$^c$, Marco Visentini-Scarzanella$^d$, Arianna Menciassi$^b$, Paolo Dario$^b$, Anastasios Koulaouzidis$^e$, Alberto Arezzo$^f$, David J Hawkes$^a$,and Danail Stoyanov$^a$}

\address{$^a$ Centre for Medical Image Computing, University College London, London, UK\\
E-mail: patrick.brandao.15@ucl.ac.uk}

\address{$^b$The BioRobotics Institute, Scuola Superiore Sant'Anna, Pisa, Italy.}

\address{$^c$Computer Science Department at Universitat Autònoma de Barcelona, Barcelona, Spain}

\address{$^d$Multimedia Laboratory Toshiba Corporate Research and Development Center, Kawasaki, Japan.}

\address{$^e$UEndoscopy Unit, The Royal Infirmary of Edinburgh, Edinburgh, UK}

\address{$^f$Department of Surgical Sciences, University of Turin, Turin, Italy.}

\maketitle

\begin{abstract}

Early diagnosis is essential for the successful treatment of bowel cancers including colorectal cancer (CRC) and capsule endoscopic imaging with robotic actuation can be a valuable diagnostic tool when combined with automated image analysis. We present a deep learning rooted detection and segmentation framework for recognizing lesions in colonoscopy and capsule endoscopy images. We restructure established convolution architectures, such as VGG and ResNets, by converting them into fully-connected convolution networks (FCNs), fine-tune them and study their capabilities for polyp segmentation and detection. We additionally use Shape-from-Shading (SfS) to recover depth and provide a richer representation of the tissue's structure in colonoscopy images. Depth is incorporated into our network models as an additional input channel to the RGB information and we demonstrate that the resulting network yields improved performance. Our networks are tested on publicly available datasets and the most accurate segmentation model achieved a mean segmentation IU of 47.78\% and 56.95\% on the ETIS-Larib and CVC-Colon datasets, respectively. For polyp detection, the top performing models we propose surpass the current state of the art with detection recalls superior to 90\% for all datasets tested. To our knowledge, we present the first work to use FCNs for polyp segmentation in addition to proposing a novel combination of SfS and RGB that boosts performance.
\end{abstract}


\begin{multicols}{2}
\section{Introduction}

Colorectal cancer (CRC) is the most frequent pathology of the gastrointestinal tract and accounts for nearly 10\% of all forms of cancer.  When the disease reaches an advanced stage, the 5-year survival rate of CRC patients is lower than 7\%, but early diagnosis with successful treatment can dramatically increase this figure to more than 90\% \cite{c1}. Conventional colonoscopy is the reference standard for CRC screening and diagnosis. It provides direct visualization of the inner surface of the colon for acquiring biopsies and performing therapeutic procedures on early stage neoplastic lesions. Wireless and tethered endoscopic capsules and robotic endoscopes have recently been reported to overcome the pain and discomfort associated with conventional colonoscopy \cite{c1}. Despite assisted articulation or the use of wireless capsules, the success of the exam still mostly depends on the operator's skills both for dexterous maneuvering of the camera and for ensuring full exploration of the colon wall \cite{c1}. With wireless capsules the problem is more focused on the processing and analysis time of the long and often occluded video signal. As a result, even experienced endoscopists can miss polyps during examination \cite{c2}. To overcome this, computer-aided diagnostic (CAD) systems can be used to support clinical decisions with automated polyp detection algorithms.

Despite significant progress in recent years, CAD polyp detection is still far away from routine clinical use. The problem is challenging because there is large variety in the size, shape, colour and textures of colon polyps, which alongside the presence of specular reflections, endoluminal folds and blood vessels in colonoscopy recordings, impedes detection accuracy and induces a significant number of false detections. Early CAD polyp detection methods used texture and colour as the main features \cite{c3,c4}. More recently, shape information has been incorporated to increase feature complexity and discriminating power and provide improved detection models \cite{c5,c6}. 

We build onto recent advances in computer vision and incorporate shape information to build a system based on a convolutional neural network (CNN) framework that can accurately detect and segment polyps and high-risk regions in colonoscopy images, while minimizing false detections. We combine state-of-the-art deep learning techniques designed for generic object detection and adapt them to endoscopic images. To our knowledge, this is the first work to fine-tune cutting edge CNN architectures, such as VGG and ResNets, for polyp segmentation. To increase detection accuracy, we also extracted a relative measure of depth from the colonoscopy images, using Shape-from-shading (SfS), and input it as an additional feature to the networks to enhance polyp representation. Our study also includes investigation into the influence of batch normalization in training convergence.

\section{Related work}

Propelled by large scale datasets such as ImageNet \cite{c8} and Pascal VOC \cite{c9}, deep learning approaches are the current paradigm for many computer vision problems, often surpassing by some way traditional methods for classification, segmentation, detection and tracking. Since the first deep CNN won the Imagenet Large Scale Visual Recognition Competition (ILSVRC) classification challenge \cite{c12} continuous breakthroughs in CNN architectures, training optimization and fast Graphic Processing Units (GPUs) \cite{c13} have led to continuous performance improvements. Studying the influence of network depth in large-scale image recognition resulted in the powerful 16 and 19 layer VGG network \cite{c14}. In the same year, the concept of "inception architecture" was developed \cite{c15} and the GoogLeNet won the Imagenet challenge. The current leader in results successfully trained a network (ResNet) 8-times deeper than VGG by using a residual learning framework \cite{c16}.  Several CNNs \cite{c21,review1,review2} were also used in semantic segmentation achieving impressive results. 
Along with training on RGB data, image depth has also been used to provide a richer scene representation, in detection and segmentation tasks using a conditional random field (CRF) model \cite{rgbd_silberman_2011}. Depth can also be used directly as a fourth input channel and achieve improved results comparing to RGB-only training \cite{rgbd_couprie_2013} and can improve the performance of the fully convolutional version of VGG by performing training with RGB-D images \cite{c21,review3}.

Large scale polyp datasets, such as CVC-ColonDB \cite{c18}, that recently became publicly available, facilitated the development of deep learning techniques for polyp detection in endoscopic images. Zhu \textit{et al.} fed features extracted through a CNN, into a Support Vector Machine (SVM) to identify lesion areas in the colon \cite{c19}. On the other hand, Tajbakhsh \textit{et al.} used a variety of pre-calculated features as inputs in a set of CNNs to classify polyp candidates \cite{c20}. However, these methods are tested in different datasets, which limits comparison between the reported performance. In 2015, the MICCAI sub-challenge on automatic polyp detection provided a set of guidelines and datasets that allowed direct comparison between different methods. The best results were achieved by models learned end-to-end. The OUS participation achieved a detection precision and recall of 69.7\% and 63.0\%, respectively, in the ETIS-Larib dataset by slightly modifying the AlexNet model. The CUMED submission used a fully convolution network (FCN), initialized from models trained in large scale datasets to achieve the best overall performance, with a 72.3\% detection precision and 69.2\% recall in the same dataset \cite{c25}. 

Despite the clear advantages that CNN offer, defining and training a CNN architecture from the beginning is not a trivial task. The training process typically requires an excessive amount of labeled data, something that is generally difficult to be obtained for medical images. In addition, selecting the architecture that provides the best compromise between convergence speed and inference ability is both challenging and time consuming. Finally, potential issues of convergence and overfitting often require significant tuning of the learning parameters \cite{c11}. Alternatively, it has been suggested \cite{c11} that fine-tuning deep-learned architectures for specific tasks (e.g. image detection and segmentation), provides better performance than designing and training a new network model. Such an approach can be  applied in tasks involving medical images, despite being substantially different from the natural images used in training the initial network, by using a smaller set of images to re-train (fine-tune) a pre-existing network. We adhere to this strategy and present a number of established CNNs which after appropriate fine-tuning are optimized for detecting and segmenting polyps in colonoscopy recordings.

We are expanding on the work developed in \cite{brandao2017fully} ,
where we segmented polyps using the same principles used by \textit{Long et al.}\cite{c21} to semantically segment objects in natural scene images. First, we exploit the use of residual networks \cite{c16} which significantly outperformed previous models in large scale recognition challenges. We also evaluate the effect of using depth estimations as a extra input feature for the best models. Finally, to guarantee generality, all models are evaluated in an extra publicly available dataset, the CVC-Colon  \cite{c26}.

\section{Proposed Methods}
We develop polyp segmentation CNN models, by utilizing well-known architectures initialized with weights obtained from pre-training, in large image-datasets and then fine-tune them with labeled colonoscopy images from publicly available databases (ETIS-Larib, CVC-ColonDB etc). The models we explore have optimized architectures and learned feature extractors that are capable of solving complex object detection tasks. For performing polyp segmentation instead of simple detection, we opted to convert the pre-trained CNNs into FCNs by adding deconvolution layers in order to obtain a pixel classification map as the output. Moreover, we applied an efficient SfS technique to recover depth from colonoscopy images and provide it as an additional, to the standard RGB channels, parameter in the formulation of our network models.
 
\subsection{CNN and FCN Basis}
Irrespective of the architecture, CNNs always integrate three basic components: convolution, activation function and pooling operation layers. They operate on local inputs, depending only on relative spatial coordinates. An example of  of the effect of convolution and activation on a colonoscopy image is depicted in Fig. \ref{fig:batchnorm}.

\begin{figurehere}
\centering
  \includegraphics[height=6cm,keepaspectratio]{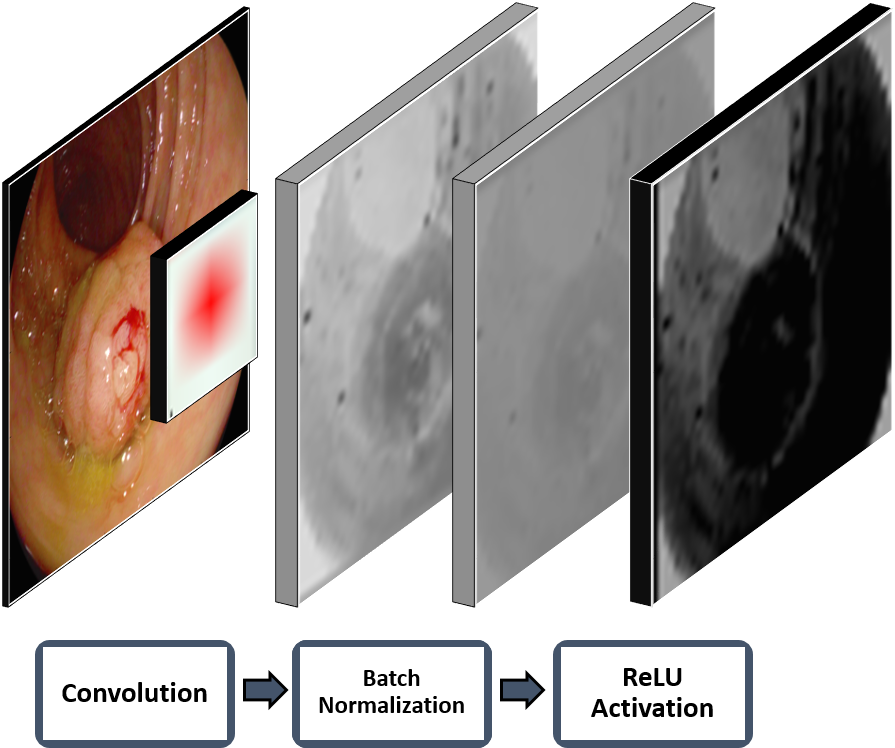}
  \caption{The basic CNN operations on a single CNN neuron from the first layer of the FCN-VGG with batch normalization. Image sequence left to right: input image, receptive field, convolution results, normalized image, and ReLU activated image.}
  \label{fig:batchnorm}
\end{figurehere}

Considering $x^{k}_{i,j}$ as the input data vector at location $(i,j)$ in layer $k$, the input of the following layer, $x_{i,j}^{k+1}$, is computed by 
\begin{equation}
x_{i,j}^{k+1}=f_{S}^k ({x_{si+\delta_{i},sj+\delta_{j}}^k }  ,0 \leq \delta_{i},\delta_{j} \leq w). \label{eq1}
\end{equation}
where $S$ is the stride or subsample factor and $f^k$ represents the type of operation of the layer $k$. In classification CNNs, the network ends with one or more fully connected layers that produce non-spatial outputs \cite{c21}. A loss function $l$ compares the prediction outputs of the last layer $f^K$ to the desired result $y$  as:
\begin{equation}
l(x^K,y)=\sum_{i,j}l (x_{i,j}^K,y_{i,j} ). \label{eq2}
\end{equation}
Using the chain rule, the gradient of the loss is back-propagated throughout the network and the parameters of all layers are updated via stochastic gradient descend (SGD) \cite{c22}. 


Traditional CNN architectures, such as AlexNet and VGG, are used for classification problems, which mean that take an input image and output a single classification score for all the possible classes. To obtain pixel wise segmentation, these networks need to be converted to a fully convolutional network. 
A fully connected layer can be viewed as a convolution layer where the kernel has the same dimensions as the input. By replacing these with convolutions, it is possible to convert traditional classification networks into FCNs that take inputs of any size and output coarse classification maps. While the resulting maps are equivalent to the processing of the individual patches on the original network, the computational cost is highly amortized by the inherent efficiency of convolution. Even though the output maps can yield any size, these are typically reduced by subsampling within the network \cite{c21}. To connect these coarse outputs to dense pixels, an interpolation strategy needs to be used.

Convolution is a linear operation, and as such, it can be expressed in a matrix multiplication form. Assuming $\Omega$ as the map of size $W\times H$ to be convoluted by the kernel $\theta$ of size $W'\times H'$ with a stride $S$, the convolution operation can be expressed as

\begin{equation}
vec(\psi)=C vec(\Omega). \label{eq3}
\end{equation}
where $vec(\Omega)$ represents $\Omega$ flattened to a $WH$ dimensional vector, $vec(\psi)$ is a vector with size $D=\big(\frac{W-W'}{S}+1\big) \times \big(\frac{H-H'}{S}+1 \big)$  and $C$ is a sparse matrix of size $D\times WH$, where the non-zero elements are elements of $K$. The vector $vec(\psi)$ can be later reshaped to a $\big(\frac{W-W'}{S}+1\big) \times \big(\frac{H-H'}{S}+1 \big)$  convoluted map [14]. During CNN training, the loss $\psi_l$ is backward passed to the lower level layers by convolution transpose
\begin{equation}
vec(\Omega_l)=C^T  vec(\psi_l). \label{eq4}	
\end{equation}

where $\Omega_l$ and $\psi_l$ have the same dimensions of $\Omega$ and $\psi$ in the forward pass, and  $\Omega_l$ connectivity pattern is compatible with $C$ by construction \cite{c23}.

If $S>1$, convolution implements a subsample operation. Intuitively, transpose convolution is a way to upsample the input by a factor of S. Following this principle, by simply reversing forward and backward pass operations, it is possible to implement in-network upsampling. The transpose convolution layer, also known as deconvolution layer, does not need to have a fixed filter (doing bilinear interpolation, for example) but can also be learned and adjusted during training. This provides very fast and effective upsampling used to structure efficient FCNs, capable of achieving state-of-the-art results in semantic segmentation \cite{c21}.

\subsection{Batch normalization}

SGD iteratively estimates the global gradient of the loss by using a limited set of samples. Changes in distribution of the inputs hinders convergence, as the parameters of each layer need to adapt to a new distribution. This slows down training by demanding lower learning rates and careful parameter initialization. In deep networks, this effect is intensified because small changes in the parameters are greatly amplified throughout the network, as the inputs of each layer are affected by parameters of all preceding layers \cite{c24}.

To overcome this, we evaluate the incorporation of batch normalization into standard CNN architectures to compensate batch distribution changes. Batch normalization layers perform in-network normalization by linearly transforming each training mini-batch to have zero mean and unit variance. This technique has proved to yield improved results on classification tasks using considerable less training iterations \cite{c24}. An example of the batch normalization process is illustrated in Fig. \ref{fig:batchnorm}.

\subsection{Shape-from-Shading}
The increased detection performance of CNNs by incorporating depth information, motivated us to employ a SfS technique \cite{sfs_ahmed_a_2007} to extract depth from colonoscopy images and include it in the formulation of our models. SfS aims to recover the 3D shape of an object by analyzing the illumination variation across the image. Subsequently, SfS is suitable for approximating depth in colonoscopy recordings with a monocular view without requiring stereo or multi-view matching \cite{mtv} and structure from motion estimation \cite{strmotion}. While limited to only relative depth, SfS does not require texture assumptions like shape-from-texture \cite{shapetexture} techniques and is useful for extracting geometric information easily from existing clinical colonoscopy systems.

The majority of SfS approaches \cite{sfs_ahmed_a_2007, sfs_ahmed_b_2007, sfs_zhang_2009} assume a light source either coinciding with the optical center or infinitely far away from the scene. These conditions are unrealistic in the case of colonoscopy even though the light source and the camera are both at the tip of the instrument. This is because despite the small distance between the camera and the light, the observed tissue is also very close and highly dependent on small illumination changes. To overcome this limitation, we use a method which approximates the position of the lightsource at the tip of the endoscope and uses the position directly in the SfS problem formulation \cite{sfs_stoyanov_2012}. 

Assuming the camera positioned at $\mathbf{y} = (a, b, c)$ and a set $\mathcal{A}$ of all image points $\mathbf{x} = (x, y)$, the SfS problem is formulated as a Hamiltonian-Jacobi non-linear Partial Differential Equation (PDE),
\begin{equation}
    H(\mathbf{x}, \nabla v)=\frac{1}{\rho}I(\mathbf{x})
    \sqrt{(v_x^2 + v_y^2) + J(\mathbf{x}, \nabla v)^2}
    \cdot
    Q(\mathbf{x})^{\frac{3}{2}},\label{eq5}
\end{equation}
where
\begin{equation}
Q(\mathbf{x}) = (x + a)^2 + (y + b)^2 + (f + c)^2,
\end{equation}
\begin{equation}
J(\mathbf{x}, \nabla v)=\frac{\left(x + a\right)v_x + \left(y + b\right)v_y + 1}{f + c},
\end{equation}
\begin{equation}
\textbf{I}(x) = \rho \frac{I.\mathbf{n}}{r^2},
\end{equation}
\begin{equation}
v = \log d(x) . \label{eq9}
\end{equation}
$d(\mathbf{x})$ is the depth of point $\mathbf{x}$, $v_x$ and $v_y$ are the spatial derivatives of $v$, $\rho$ is the surface albedo and $\mathbf{I}$ and $\mathbf{n}$ are the light and surface normal vectors, respectively. The Lax-Friedrichs method \cite{sfs_lax_friedrichs_2004} is used to solve the resulting PDE and a specular highlight triangulation method is used to solve for the unknown albedo. An example of depth extraction from a single colonoscopy image can be seen in Fig. 
\ref{fig:sfs}.

\begin{figurehere}
 \centering
 \minipage{0.14\textwidth}
 \centering
    \includegraphics[width=1.0\textwidth]{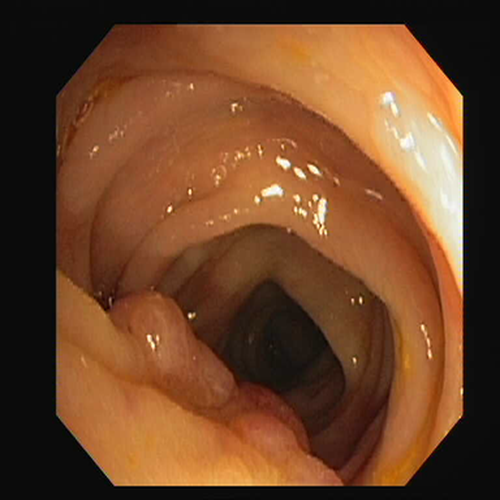}\           (a)
\endminipage \hfill
 \minipage{0.14\textwidth}
    \centering
    \includegraphics[width=1.0\textwidth]{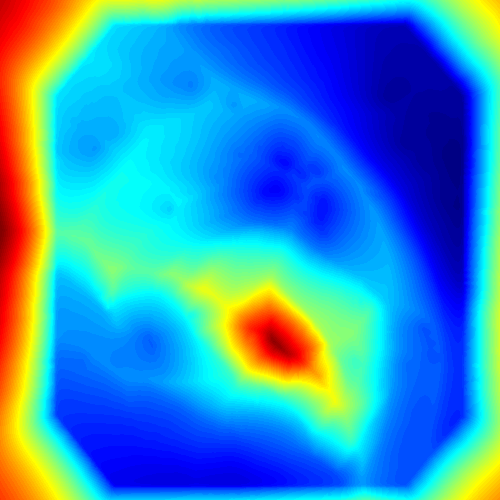}\           (b)
\endminipage
 \minipage{0.19\textwidth}
    \centering
    \includegraphics[width=1.0\textwidth]{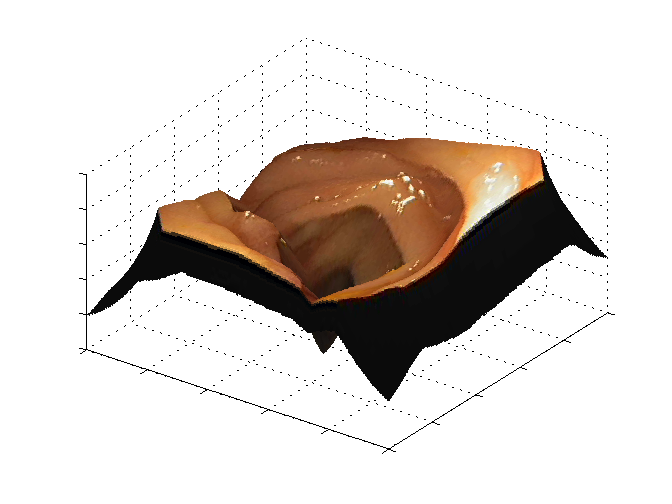}\           (c)
\endminipage
\caption{SfS method employed. (a) image from the CVC-ClinicDB dataset; (b) depth estimation from SfS; (c) 3D surface recovered from SfS depth.}
\label{fig:sfs}
\end{figurehere}

\begin{figure*}[!ht]
\centering
  \includegraphics[width=1.0\textwidth,keepaspectratio]{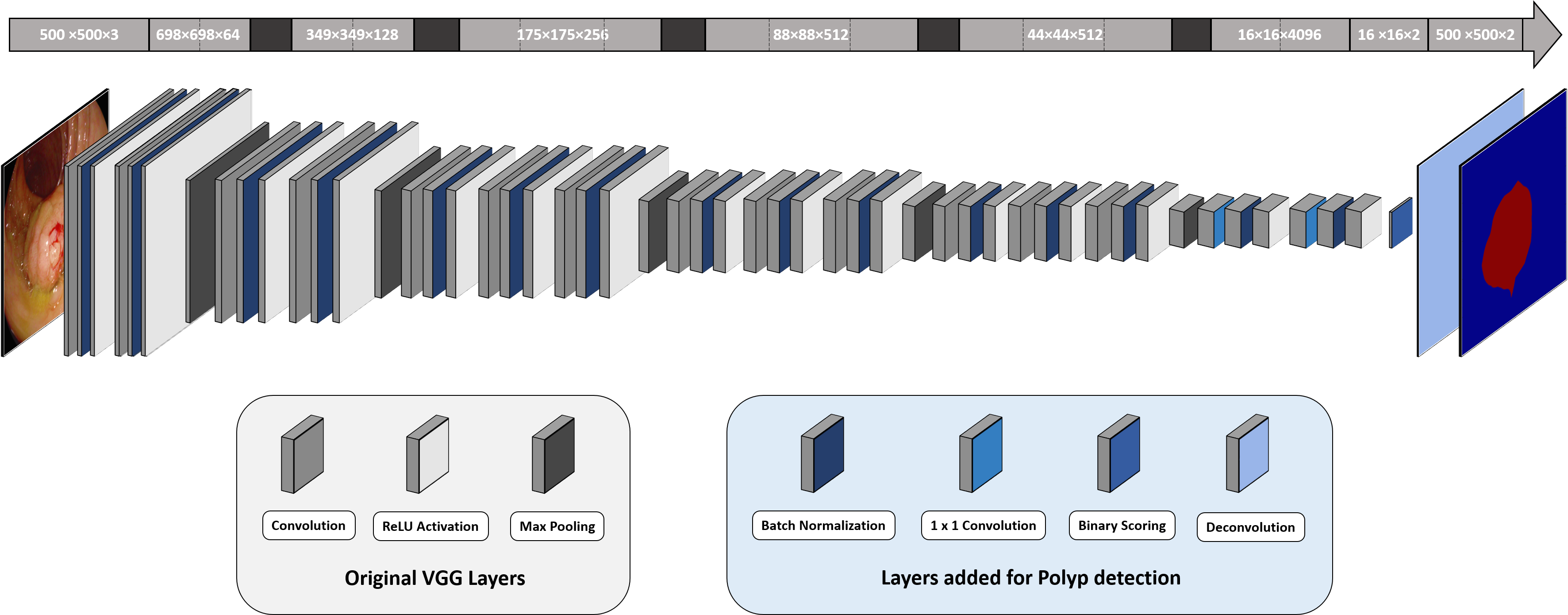}
  \caption{  Illustration of the proposed BN-FCN-VGG architecture with batch normalization for a $500\times 500$ size image. The values on the top array represent the output size of each layer underneath.  The fully connected and scoring layers of the original VGG were removed. Grey coloured layers were loaded from the original model while blue coloured layers were added or modified for polyp segmentation.}
  \label{fig:vgg}
\end{figure*}

\subsection{Proposed architectures for polyp detection}
We investigated several state-of-the-art convolution architectures and adapt them through fine-tuning, for polyp detection. Specifically, we test six different architectures: AlexNet \cite{c12}, GoogLeNet \cite{c15}, VVG \cite{c14} and three version of the ResNets architecture with 50, 101 and 152 layers of depth \cite{c16}. The AlexNet and VGG are converted into FCNs (FCN-AlexNet and FCN-VGG) by discarding the two fully connected layers and replacing them with $1\times 1$ convolution layers with  the same 4096 dimensions of the fully connected layers. The final scoring layers were also replaced with a 2D, $1\times 1$ convolution to produce the background and polyp pixel classification maps as the output. The conversion of GoogLeNet and ResNets into FCNs only requires the replacement of the scoring layers with a 2D convolution. We also increased the resolution of the output coarse map by discarding the final averaging layer. ResNets already incorporate batch normalization, while to apply this to the remaining networks, we added a regularization operation between every convolution and activation layer, as illustrated in Fig. \ref{fig:batchnorm}. Every network is finalized with a deconvolution layer with stride $S=32$ and a kernel of size $W'= H'= 64$ , responsible for upsampling the coarse output to a dense scoring map with the same dimensions as the input. Even though the CNNs output a coarse segmentation map, a single deconvolution layer can accurately upsample the blob-like structures of most polyps. We verified that adding extra deconvolution layers from the finer levels of the models did not improve the results.    An example the proposed fully connected version of VGG with batch normalization (BN-FCN-VGG) is illustrated in Fig. \ref{fig:vgg}.  

\subsection{Implementation and training details}
Developed networks were optimized by SGD with a 0.99 momentum and all layers were updated by back-propagation. Classes probabilities are calculated with Softmax function and cross-entropy was used as the loss function and the learning rate for convolution bias was doubled. In GoogLeNet, the two deeper loss function were discarded and only the last one was used for fine-tuning. For FCN-Alexnet and FCN-VGG, the convolution filters were initialized by copying weights from public available models trained on the PASCAL segmentation dataset. Because no trained segmentation models are publicly available,  the fully convolutional GoogLeNet (FCN-GoogLeNet) and the three fully convolution ResNets (FCN-ResNet)  were initialized by loading classification models trained on the Imagenet dataset. New convolution layers were zero-initialized and the learning rate of scoring layers was increased by a factor of 10. We fine-tune the networks with the highest fixed learning rate that did not cause loss divergence. For FCN-GoogLeNet this corresponds to a learning rate of $10^{-12}$, while all other FCNs were optimized with a learning rate of $10^{-10}$. Convergence was achieved after 30K iterations for FCN-ResNet-51 and FCN-ResNet-101, 40K for FCN-GoogLeNet, 50K for FCN-VGG and FCN-ResNet-151 and 120K for FCN-AlexNet.

Images were  resized to $500 \times 500$ and random flipping was used for data augmentation during training. Non residual FCNs were trained with a random single image per batch. All FCN-ResNets were trained with $224\times224$ patches randomly sampled from the training images. This allowed to increase the batch size even with limited memory resources. The same type of sampling was performed during training of the batch normalization versions (BN-FCNs) of the non residual networks. Batch sizes of 20 were used for BN-FCN-AlexNet, BN-FCN-GoogLeNet and FCN-ResNet-51. Due to memory constrains, smaller batch sizes were used for other FCNs: 16 for FCN-ResNet-101, 8 for FCN-ResNet-151 and 5 for BN-FCN-VGG. When training networks with depth (D-FCN), the SfS values are concatenated to the RGB channels to create a new 4-channel input. A new channel is added to every first layer convolution filter and it is initialized by averaging the values of the other filter dimensions. The learning rate of this layer is increased by a factor of 10.  All models were trained and tested using the Caffe \cite{c30} software library in a single NVIDIA Tesla K40 GPU.

\begin{figure*}
\centering
  \includegraphics[width=1.0\textwidth,keepaspectratio]{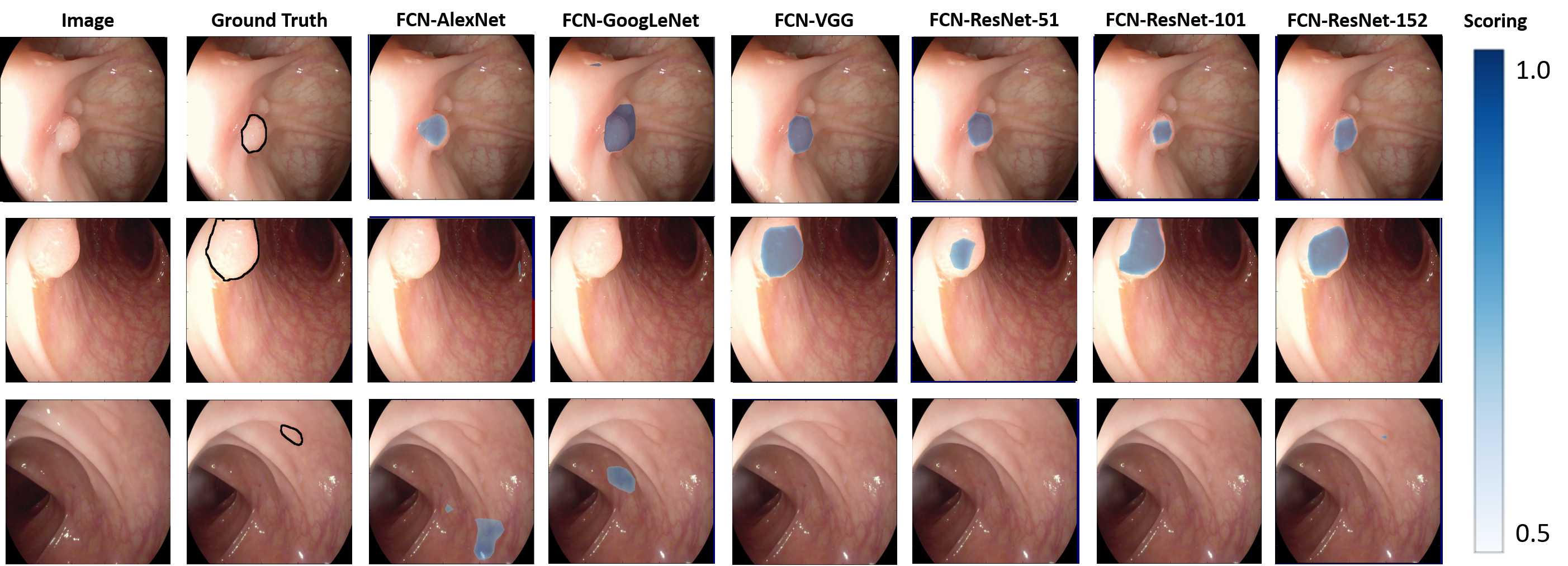}
  \caption{Example of three different scored segmentations produced by the six proposed FCN networks. The colorbar defines the scoring probability of each pixel to belong to the polyp class. Third image results are best viewed in colour electronically because the FCN-ResNet-152 detection is very small.}
  \label{fig:res1}
\end{figure*}

\begin{table*}[ht]
\centering
\tbl{Segmentation and detection precision (prec) and recall (rec) in \% obtained by the proposed FCNs. Mean interception over union (IU) is also presented for segmentation. The best result for each metric is highlighted \label{tab:res1}}{

\begin{tabular}{ccccccccccc}
\hline
                                                                      & \multicolumn{5}{c}{\textbf{ETIS-Larib}}                                                                                                                                             & \multicolumn{5}{c}{\textbf{CVC-ColonDB}}                                                                                                                                             \\ 
                                                                      & \multicolumn{3}{c}{\textbf{Segmentation}}                                                                 & \multicolumn{2}{c}{\textbf{Detection}}                                 & \multicolumn{3}{c}{\textbf{Segmentation}}                                                                  & \multicolumn{2}{c}{\textbf{Detection}}                                 \\ 
\multicolumn{1}{l}{}                                                 & \multicolumn{1}{c}{\textbf{Prec}} & \multicolumn{1}{c}{\textbf{Rec}} & \multicolumn{1}{c}{\textbf{IU}} & \multicolumn{1}{c}{\textbf{Prec}} & \multicolumn{1}{c}{\textbf{Rec}} & \multicolumn{1}{c}{\textbf{Prec}} & \multicolumn{1}{c}{\textbf{Prec}} & \multicolumn{1}{c}{\textbf{IU}} & \multicolumn{1}{c}{\textbf{Prec}} & \multicolumn{1}{c}{\textbf{Rec}} \\ \hline

\multicolumn{1}{c}{\textbf{FCN-AlexNet}}    &27.87	& 35.54 &	15.7	 & 44.08 &	63.78 &	40.3 &	20.71 &	15.77 &	45.29 &	54.68
                         \\ 
\multicolumn{1}{c}{\textbf{FCN-GoogLeNet}}                          & 25.83	 & 29.82 &	12.29 &	41.85 &	62.76 &	37.46 &	12.93 &	12.71 &	42.26 &	45.25 
                           \\

\multicolumn{1}{c}{\textbf{FCN-VGG}}        & \textbf{70.23} &	\textbf{54.2} &	\textbf{44.06} &	73.61 &	86.31 &	\textbf{76.06} &	\textbf{60.46}&	\textbf{54.01} &	79.57 &	86.01
                          \\ 

\multicolumn{1}{c}{\textbf{FCN-ResNet-50}}                          & 55.75                              & 23.43                             & 19.72                                 & 73.84                              & 76.53                              & 67.76                              & 25.64                              & 22.74                                 & 82.89                              & 82.38                              \\

\multicolumn{1}{c}{\textbf{FCN-ResNet-101}} & 63.26                              & 53.88                              & 41.35                                 & 75.32                              &\textbf{ 91.66}                              & 73.85                              & 50.73                              & 46.23                                 & \textbf{83.70 }                             & 88.20                              \\ 
\multicolumn{1}{c}{\textbf{FCN-ResNet-152}}                         & 65.26                              & 38.24                             & 33.19                                 & \textbf{79.42 }                             & 89.75                              & 72.85                              & 50.72                              & 43.28                                 & 82.08                              & \textbf{93.27}                              \\ \hline
\end{tabular}
}
\end{table*}

\section{Experiments and Results}

We used the public datasets from the MICCAI 2015 polyp detection challenge \cite{c25}. For comparison purposes, we divided the dataset for training and testing as it was suggested in the MICCAI challenge guidelines: CVC-CLINIC and ASU-Mayo for training and ETIS-Larib for testing  . Furthermore, we also report results from a second public available dataset (CVC-ColonDB) \cite{c26}. The datasets were obtained with different imaging systems and contain manual segmentations of every detected polyp. We specifically used the following grouping of images for training (fine-tuning) and testing:

\begin{itemlist}
\item {CVC-CLINIC: 612 SD training frames with at least one polyp each;}
\item {ETIS-Larib: 196 HD testing frames with at least one polyp each;}
\item {ASU-Mayo: 36 small SD and HD videos sequences, divided into training frames with and without polyps.}
\item {CVC-ColonDB: 379 testing frames from 15 different colonoscopy sequences with at least one polyp each.}
\end{itemlist}

In total, the MICCAI challenge training data has 19514 frames from CVC-CLINIC and ASU-Mayo datasets. However, only 4664 of these corresponds to images with polyps. We verified that networks trained with the full dataset did not converge satisfactory in practical time and performance was substantially worst. To avoid this, we fine-tuned all proposed FCNs only with images containing polyps, achieving better performance. 

The developed FCNs were formulated to produce dense pixel-wise polyp segmentations. As such, we report results using three common segmentation evaluation metrics: mean pixel precision, mean pixel recall and interception over union (IU). If a pixel of polyp is correctly classified it is counted as a true positive (TP). Every pixels segmented as polyp that fall outside of a polyp mask counts as a false positive (FP). Finally, Every polyp pixel that has not been detected counts as a false negative (FN). The evaluation metrics are calculated according Equation \ref{eq11}.

\begin{equation}
 \label{eq11}
\resizebox{0.4\textwidth}{!}{
$Prec =  \frac{TP}{TP+FP}  \qquad Rec =  \frac{TP}{TP+FN} \qquad IU=\frac{TP}{TP+FP+FN}.$
}
\end{equation}

Since, in the MICCAI challenge results are reported in terms of polyp detection, we also evaluate the polyp detection rate using the metrics advocated by the challenge directives \cite{c25}; detection precision and recall. If a segmented blob falls within the polyp mask it is counted as a TP. If the detected blob falls outside the ground truth mask it is a FP. Every polyp in the image that has not been detected counts as a FN. Only one TP is considered for polyp, no matter how many detections fall within the polyp mask. Detection precision and detection are calculated with the same formulas of Equation \ref{eq11}

\begin{figure*}
\centering
  \includegraphics[width=0.7\textwidth,keepaspectratio]{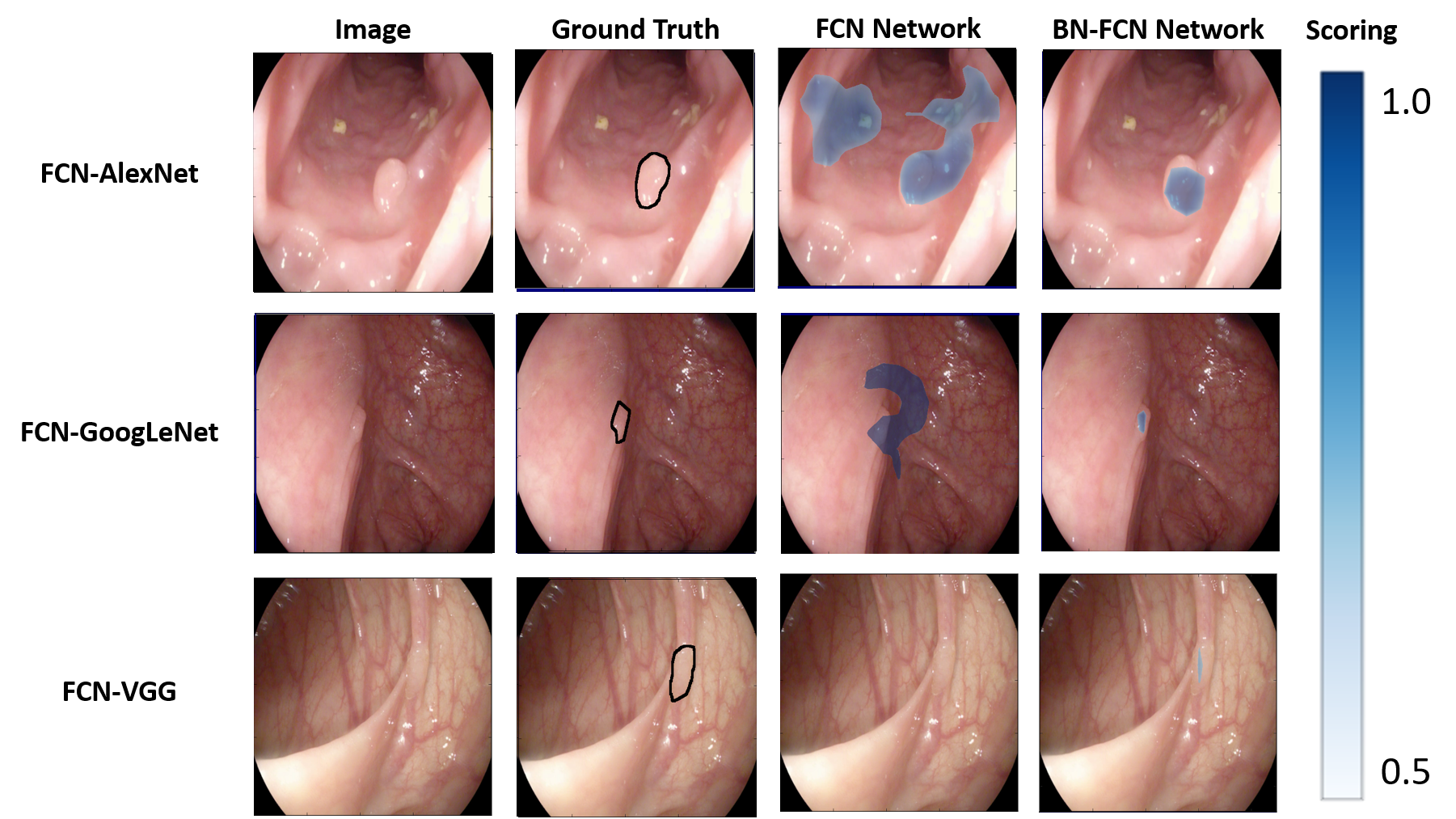}
  \caption{Segmentation comparison obtained by the three non-residual architectures with and without batch normalization. FCN-VGG results are best viewed in colour electronically the detection is very small.}
  \label{fig:batch_results}
\end{figure*}
\subsection{Results from RGB data}

\begin{table*}
\centering
\tbl{Segmentation and detection precision (prec) and recall (rec)  in \% obtained by the non residual FCNs trained with batch normalization. Mean interception over union (IU) is also presented for segmentation. Metrics improved by adding BN are highlighted in bold  \label{tab:res2}}{

\begin{tabular}{ccccccccccc}
\hline
                                                               & \multicolumn{5}{c}{\textbf{ETIS-Larib}}                                    & \multicolumn{5}{c}{\textbf{CVC-ColonDB}}                                   \\ 
                                                               & \multicolumn{3}{c}{\textbf{Segmentation}} & \multicolumn{2}{c}{\textbf{Detection}} & \multicolumn{3}{c}{\textbf{Segmentation}} & \multicolumn{2}{c}{\textbf{Detection}} \\ 
                                                               & \textbf{Prec}      & \textbf{Rec}       & \textbf{IU}        & \textbf{Prec}           & \textbf{Rec}           & \textbf{Prec}      & \textbf{Rec}       & \textbf{IU}        & \textbf{Prec}           & \textbf{Rec}           \\ \hline

\multicolumn{1}{c}{\textbf{BN-FCN-AlexNet}}   & 
\textbf{30.05} &	29.07 &\textbf{	17.41} &	38.95 & 62.76 &\textbf{	46.22} &	\textbf{28.78} &	\textbf{21.19 }&	43.69 &	\textbf{80.87  }      \\ 
\multicolumn{1}{c}{\textbf{BN-FCN-GoogLeNet}} & \textbf{49.36}        &23.85       & \textbf{20.36}        & \textbf{53.96 }            & \textbf{63.10 }          & \textbf{63.87}        & \textbf{25.92}        & \textbf{23.04 }      &   \textbf{ 62.56}          & \textbf{75.99}           \\

\multicolumn{1}{c}{\textbf{BN-FCN-VGG}}       & 56.87	&\textbf{ 66.59} &	42.32 &	56.24 &	\textbf{94.01} &	66.8 &	\textbf{61.3} &	47.18 &	61.57 &	\textbf{95.16}
    \\ \hline
\end{tabular}
}
\end{table*}

We first train every FCNs using only RGB data and compare their performance on both testing datasets. Table \ref{tab:res1} presents the segmentation and detection results for all proposed network architectures and Fig. \ref{fig:res1} illustrates representative examples of polyp segmentation for each network. FCN-ResNet-152 and FCN-ResNet-101 proved to be the best polyp detectors achieving the highest recalls in both databases. In some situations (third row in Fig. \ref{fig:res1}), FCN-ResNet-152 was the only network capable of correctly detecting the polyp, even with limited segmentation accuracy.  Both deep architectures (FCN-ResNet-101, FCN-ResNet-152) proved to be able to learn complex filters capable of 90\%detection recall in the testing datasets. FCN-ResNet-50 resulted in less accurate detections than its deeper counterparts, with approximately 10\% lower detection recall.  These observations indicate that, while more than 50 layers are essential in handling the high complexity of detecting polyps of various sizes, shapes and textures, as indicated by the improved performance of FCN-ResNet-100 against FCN-ResNet-50, the addition of even more layers in FCN-ResNet-152 does not improve the detection performance.  

In the non residual architectures, FCN-VGG outperforms the other FCNs by achieving detection recalls of 86\% in both datasets. The simpler FCN-AlexNet successfully detected 63.78\% of the ETIS-Larib polyps, and 54.68\% of the CVC-ColonDB, and resulted in a considerable amount of false positives, as exemplified by the second and third segmentations of Fig. \ref{fig:res1}. Finally, the FCN-GoogLeNet produced the worst detection performance of all networks studied. Although, GoogLeNet is a deeper architecture than the other two, this does not necessarily translate to better inference ability, as the network is notoriously hard to optimize. Without the two loss functions that allow better convergence in the deeper modules, the network is not able to completely adapt to the new learning problem.

\begin{figure*}
\centering
 \includegraphics[width=1.0\textwidth,keepaspectratio]{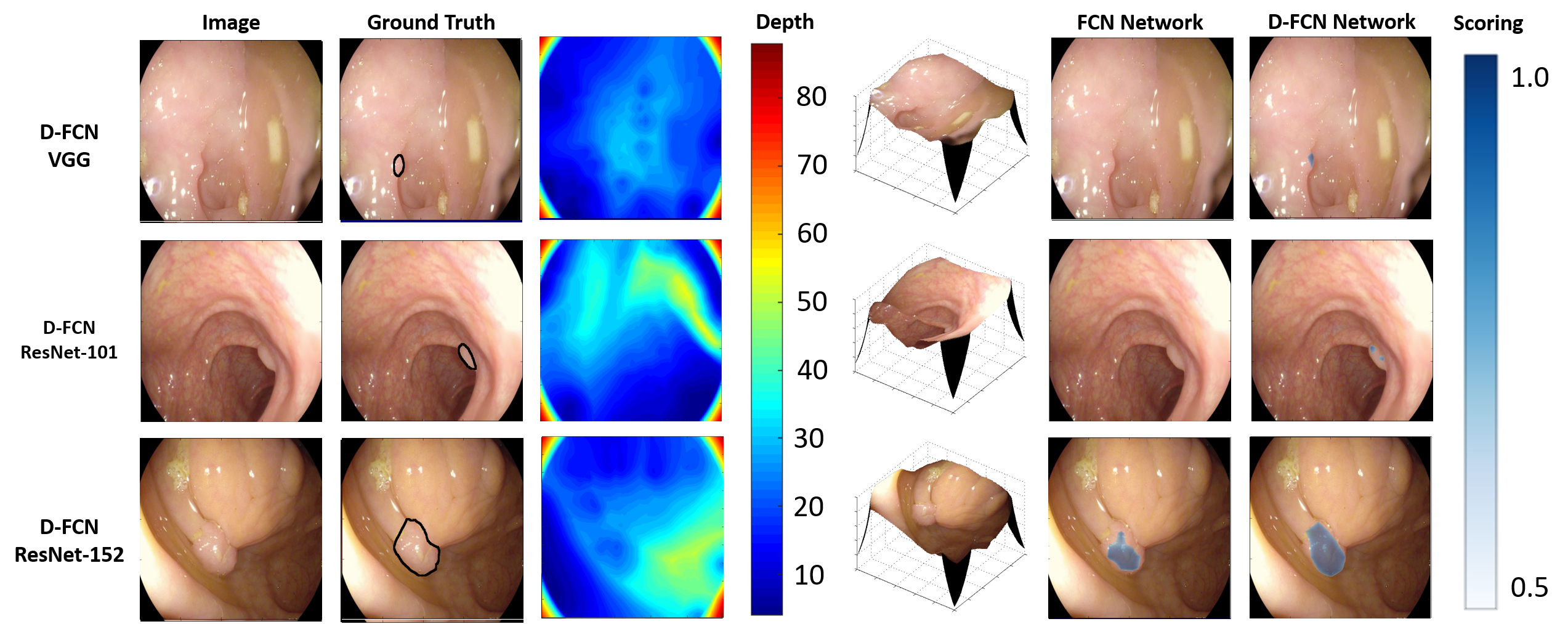}
  \caption{Comparison between segmentations obtained by the three top-performing architectures trained with and without depth.}
  \label{fig:rgbd_results}
\end{figure*}

\begin{table*}
\centering
\tbl{Segmentation and detection precision (prec) and recall (rec)  in \% obtained by the three FCNs with the best performance trained with RGB-D data (D-FCNs). Mean interception over union (IU) is also presented for segmentation. Metrics improved by adding depth information are highlighted in bold \label{tab:res3}}{

\begin{tabular}{ccccccccccc}
\hline
                                                               & \multicolumn{5}{c}{\textbf{ETIS-Larib}}                                    & \multicolumn{5}{c}{\textbf{CVC-ColonDB}}                                   \\
                                                               & \multicolumn{3}{c}{\textbf{Segmentation}} & \multicolumn{2}{c}{\textbf{Detection}} & \multicolumn{3}{c}{\textbf{Segmentation}} & \multicolumn{2}{c}{\textbf{Detection}} \\ 
                                                               & \textbf{Prec}      & \textbf{Rec}       & \textbf{IU}        & \textbf{Prec}           & \textbf{Rec}           & \textbf{Prec}      & \textbf{Rec}       & \textbf{IU}        & \textbf{Prec}           & \textbf{Rec}           \\ \hline

\multicolumn{1}{c}{\textbf{D-FCN-VGG}}   & 
68.68	& \textbf{62.16} &	\textbf{47.78} & 73.32 &	\textbf{88.01} &		74.94 &	\textbf{68.02} & \textbf{56.95} & 	76.85 &	\textbf{91.03}	    \\
\multicolumn{1}{c}{\textbf{D-FCN-ResNet-101}} & 55.63 & 	\textbf{61.11} &	40.99 &	70.62 &\textbf{	95.83} &	\textbf{74.31} &	\textbf{58.15} &	\textbf{49.65} &	83.17 &	\textbf{90.47}
          \\ 

\multicolumn{1}{c}{\textbf{D-FCN-ResNet-152}}       & \textbf{67.66} &	\textbf{39.78} &	\textbf{33.77} &	77.95 &	\textbf{90.2} &	71.16 &	47.95 &	41.32 &	80.78 &	92.5

    \\ \hline
\end{tabular}
}
\end{table*}

In terms of the segmentation results, FCN-VGG outperformed all other networks with an IU of 44.06\% and 54.01\% for ETIS-Larib and CVC-ColonDB datasets, respectively. Subsequently, even though FCN-VGG detects a smaller number of polyps, the overall quality of the segmentation it provides is superior to other networks. An example of this is depicted in the second polyp of Fig. \ref{fig:res1}. Similar levels of segmentation quality were achieved by FCN-ResNet-101, which indicates that it learned more general filters than its deeper residual counterpart (FCN-ResNet-151). Finally, similar to the detection results, FCN-AlexNet and FCN-GoogLeNet achieved the worst performance in segmenting the polyps.

As far as we know, our method is the first to produce dense polyp segmentations, which only allow comparison with other algorithms with the use of detection metrics. The current state of the art was set by the top deep learning method in the 2015 MICCAI polyp detection challenge (OUS), which achieved 73.3\% detection precision and 69.2\% recall in the ETIS-Larib dataset \cite{c25}. As seen in Table \ref{tab:res1}, four of our models (FCN-VGG and all three FCN-ResNets) surpass this results, with improvements in precision and increases in recall as high as 20\%. The OUS methodology was not made publicly available yet, so direct comparison is not possible. However, the huge difference in accuracy shows how important proven CNN architectures and a good initialization are to achieved a better solution.

\subsection{Adding batch normalization}
Batch normalization is not implemented in the original AlexNet, VGG and GoogLeNet. We investigate the influence of adding batch normalization in these FCNs and list the results in Table \ref{tab:res2}. During training of the batch-normalized (BN-FCNs) versions, we decreased the learning rate by a factor of 100 and convergence was achieved after 30K iterations for all BN-FCNs. Due to memory limitations, relatively small batch sizes were used. BN-FCN-AlexNet resulted in a slight increase in IU segmentation and detection recall for ETIS-Larib, while in CVC-ColonDB, every single evaluation metric was improved, especially for detection, where the recall increased by more than 25\%. Similar improvements in segmentation IU and detection accuracy, are observed with BN-FCN-GoogLeNet for both datasets. Examples of improved segmentations are illustrated in Fig. \ref{fig:batch_results} for all three non-residual networks. Batch normalization enabled BN-FCN-VGG to increase the amount of polyps detected, with recalls higher than 94\% for both datasets. However, this was accompanied with an decrease in precision.  The third row in Fig. \ref{fig:batch_results} shows an example of a polyp being misdetected without batch normalization (FCN-VGG) while being successfully recovered in the batch-normalized version (BN-FCN-VGG).

\subsection{Results from RGB-D data}
To evaluate the addition of SfS-extracted depth, as an additional feature, we restricted ourselves to the three architectures that achieved the best detection and segmentation results with RGB data; FCN-VGG, FCN-ResNet-101 and FCN-ResNet-152. The results after the inclusion of depth information are listed in Table \ref{tab:res3}.

The addition of depth information allowed D-FCN-VGG to perform slightly better than its RGB counterpart. Segmentation IU and detection recall improved approximately by 2\% for both datasets. Similar increases were verified with D-FCN-ResNet-101, elevating its detection recall to more than 95\% for the ETIS-Larib. Fig \ref{fig:rgbd_results} illustrates three examples where depth information allowed the networks to either detect a polyp that would otherwise miss (D-FCN-VGG, D-FCN-ResNets-101) or improve segmentation accuracy (D-FCN-ResNets-152). 

With the current architectures, bigger improvements are hampered by the difficulty of propagating meaningful gradients through the model. This issue is more evident in the FCN-ResNet-152, which had comparable detection performance with and without depth data. Alternative ways to incorporate depth information into the models might facilitate the learning of more meaningful RGB-D feature extractors. 

\subsection{Computation speed}
The inference speed of each network highly depends of the amount of learned parameters and the number of layers. Table \ref{tab:speed} lists the average time required for each FCN to segment a single $500\times500$ image. The addition of batch normalization and the use of depth features slows down inference, as more operations are required to produce the final segmentation maps. The VGG architecture has the highest number of receptive fields, which results in the slowest average inference speed of all networks. On the other end, FCN-AlexNet has the fastest average inference both with (136ms) and without (51ms) batch normalization.

\begin{tablehere}
\tbl{Average inference time in milliseconds (ms) for a $500\times500$ image. If applicable, average inference time is shown , without batch normalization (no BN), with batch normalization (BN) and with the inclusion of depth (Depth)\label{tab:speed}}{
\begin{tabular}{cccc}
\hline                                                      & \multicolumn{3}{c}{\textbf{Average Inference Speed (ms)}} \\
\multicolumn{1}{c}{\textbf{Networks}}                               & \textbf{no BN}    & \textbf{BN}    & \textbf{Depth}   \\ \hline
\multicolumn{1}{c}{\textbf{FCN-AlexNet}}                            & 51                & 136            & \textbf{-}            \\ 

\multicolumn{1}{c}{\textbf{FCN-GoogLeNet}}  & 60                & 193              & -                     \\ 
\multicolumn{1}{c}{\textbf{FCN-VGG}}                                & 295               & 412            & 536                    \\ 

\multicolumn{1}{c}{\textbf{FCN-ResNet-51}}  & -                 & 164            & -                     \\ 
\multicolumn{1}{c}{\textbf{FCN-ResNet-101}}                         & -                 & 206            & 265                     \\ 

\multicolumn{1}{c}{\textbf{FCN-ResNet-152}} & -                 & 319            & 387                     \\ \hline
\end{tabular}}
\end{tablehere}

\section{Conclusion}\vspace{-2ex}
We have presented a deep learning framework for automatically detecting and segmenting polyps in colonoscopy images. This is achieved by taking advantage of very rich representations available in CNNs trained on large databases which we fine-tune to perform polyp detection and adapt them, by converting them to FCNs for achieving segmentation. We compare the networks' ability to accurately detect and segment polyp structures in experiments on publicly available datasets with annotated ground truth. Obtained results suggest that the two deepest residual architectures (ResNet-101, ResNet-152) were able to cope with the complexity of polyp structures and achieve the best detection results. On the other hand, the higher number of receptive fields in the VGG, allowed the network to achieve a better overall segmentation output. These three networks achieved detection recall rates around 90\% both in the ETIS-Larib and CVC-ColonDB, considerably surpassing the state-of-the-art in polyp detection. We also introduce relative depth information, derived from SfS as an additional input channel. Results show that including depth can improve polyp representation and lead to increased detection rates and segmentation accuracy. We note thought that it is hard to propagate meaningful gradients through deep networks, thus the directly visible improvement was small. In the future, a modeling strategy that facilitates learning from depth might yield better prediction and inference rates rather than direct SfS embedding. Inference times are still a limitation for the use of CNNs in real-time CAD systems suitable for clinical practice. Strategies to speed up the inference process, such as the use of FFT convolutions, could be used for achieving minimum latency, required in clinical applications. Furthermore, a more detailed analysis of receptive field activation might allow the compression of large scale CNNs for simpler problems, providing even faster inference times.

\bibliographystyle{IEEEtran}

\end{multicols}
\end{document}